# Unleashing the Potential of Large Language Model: Zero-shot VQA for Flood Disaster Scenario

Yimin Sun

Shanghai University, yimin_sun@shu.edu.cn

Chao Wang*

Shanghai University, cwang@shu.edu.cn

Yan Peng

Shanghai University, pengyan@shu.edu.cn

Visual question answering (VQA) is a fundamental and essential AI task, and VQA-based disaster scenario understanding is a hot research topic. For instance, we can ask questions about a disaster image by the VQA model and the answer can help identify whether anyone or anything is affected by the disaster. However, previous VQA models for disaster damage assessment have some shortcomings, such as limited candidate answer space, monotonous question types, and limited answering capability of existing models. In this paper, we propose a zero-shot VQA model named **Z**ero-shot VQA for **F**lood **D**isaster **D**amage **A**ssessment (ZFDDA). It is a VQA model for damage assessment without pre-training. Also, with flood disaster as the main research object, we build a **F**reestyle **F**lood **D**isaster **I**mage **Q**uestion **A**nswering dataset (FFD-IQA) to evaluate our VQA model. This new dataset expands the question types to include free-form, multiple-choice, and yes-no questions. At the same time, we expand the size of the previous dataset to contain a total of 2,058 images and 22,422 question-meta ground truth pairs. Most importantly, our model uses well-designed chain of thought (CoT) demonstrations to unlock the potential of the large language model, allowing zero-shot VQA to show better performance in disaster scenarios. The experimental results show that the accuracy in answering complex questions is greatly improved with CoT prompts. Our study provides a research basis for subsequent research of VQA for other disaster scenarios.

CCS CONCEPTS • Computing methodologies~Artificial intelligence~Computer vision~Computer vision tasks~Scene understanding • Computing methodologies~Artificial intelligence~Natural language processing~Natural language generation

---

* Corresponding author

## 1 INTRODUCTION

Natural disasters such as earthquakes, hurricanes, and floods can cause massive destruction and casualties. On average 45,000 people die globally each year as a result of natural disasters [1]. After a natural disaster event, it is critical to quickly and accurately assess the extent of the damage to rescue. VQA is the task of answering open-ended questions based on images that require an understanding of vision, language, and common sense. VQA models allow people to ask questions in natural language from images, thereby obtaining immediate and valid information about the scene. Therefore, with the help of the VQA model, we can get valuable information about the affected area from the disaster and make decisions for rescue. Currently, some researchers use VQA for disaster scenarios. For example, SAM-VQA [2] is an attention-based VQA model for post-disaster damage assessment of remotely sensed images, and VQA-Aid [3] is a visual question-answering system integrated with UAVs, which is mainly used for post-hurricane damage assessment. Figure 1 shows an example of how VQA works in a disaster scenario.

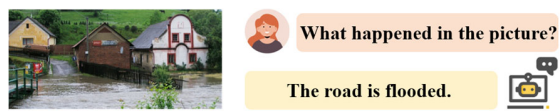

Figure 1: An overview of a disaster VQA system

Previous VQA models are mainly simple baseline models [4] (i.e., simple concatenation of image and text features) and bilinear pooling models [5]. Bilinear pooling models can also introduce an attention mechanism. These models show good visual understanding ability in VQA tasks. However, these models have the following limitations when applying VQA to disaster scenarios. **Limitation 1: Closed candidate answer space.** Most of the previous VQA models apply to datasets where all possible answers are known. However, many situations in disaster scenarios cannot be predicted in advance. The possible answers cannot be enumerated. The evaluation of disaster scenarios requires a VQA system that can generate non-fixed format and non-fixed length answers. **Limitation 2: Dataset construction still relies heavily on manual labor.** Images in previous flood disaster datasets are taken by UAVs or extracted from social media. Question-ground truth pairs in these datasets are typically annotated by professionals. When applying the VQA task to a new domain, a specialized dataset is needed to evaluate our approach. However, building the dataset manually is labor-intensive, which is a huge challenge. **Limitation 3: The potential of large language models for disaster scenarios is not effectively exploited.** Some researchers [6] build a disaster VQA system using CLIP, which is a zero-shot VQA system. However, their models do not achieve the expected results. Therefore, improving the performance of existing zero-shot VQA using more advanced techniques is also one of the issues to be investigated in this paper.

To overcome these issues mentioned, we propose a **Z**ero-shot VQA for **F**lood **D**isaster **D**amage **A**ssessment (ZFDDA), a zero-shot VQA model. As far as we know, our work is the first practice using the zero-shot VQA model as a disaster damage assessment question and answer system. Specifically, for **Limitation 1**: ZFDDA is a zero-shot VQA model. It has the following characteristics: the model can reason directly on unseen disaster datasets without any training, the model outputs answers that are variable in length and format, and the space of possible answers is infinite. For **Limitation 2**: we construct a novel disaster dataset **F**reestyle **F**lood **D**isaster **I**mage **Q**uestion **A**nswering (FFD-IQA) for evaluating our approach. We extend the question types and the size of the dataset. This dataset focuses more on vulnerable individuals and immediate disaster aid. Meanwhile, to solve the heavy reliance on manual labor in constructing new datasets, a framework for automatically constructing the dataset is proposed as well. The framework consists of three main modules: *image content extraction module*, *object matching module*, and *question generation module*. For **Limitation 3**: A well-designed CoT [7] demonstration is a key to unlocking the potential of the large language model. The introduction of the CoT improves substantially the VQA's ability to answer disaster-related questions. Currently, this study only focuses on the flood images dataset, but the model proposed by us can be directly used for any disaster scene.

Finally, we run our dataset on our model for experiments and compare it with the baseline model, on which our model expresses extremely high competitiveness. Overall, **the main contributions of our research are as follows**: (1) We propose a zero-shot VQA model ZFDDA for disaster damage assessment, and to the best of our knowledge, this is the first work that zero-shot VQA is used for disaster damage scenarios. (2) We build a novel dataset containing 2,058 flood images and 22,422 question-meta ground truth pairs. This dataset contains three types of questions: free-form, multiple-choice, and yes-no questions. In addition, we propose an automated dataset construction method to alleviate the time-consuming and labor-intensive problems caused by manually constructing the dataset. (3) We use CoT demonstration to further inspire the reasoning ability of the large language model, which makes our model perform better on the VQA task of flood disaster. Our research also provides a research basis and ideas for subsequent applications to other disaster scenarios.

## 2 RELATED WORK

### 2.1 Specific Domain VQA

VQA models have been used in both cultural and medical domains. In the cultural domain, Garcia et al. [8] proposed a VQA dataset, the AQUA dataset, dedicated to works of art and proposed a two-branch model called VIKING as a baseline. This baseline model used passages retrieved from the Knowledge Base of paintings to predict answers related to questions and paintings. Bongini et al. [9] combined the VQA and QA models. On a dataset of images about cultural heritage, the model first classified the questions and then generated accurate answers with the corresponding model. The model could help people to better understand cultural heritage. In the medical domain, Ben Abacha et al. [10] proposed VQA-Med, a medical VQA task designed to evaluate the ability of models to answer questions about medical images. This research was important for advancing VQA research in the medical field to facilitate automatic understanding and interpretation of medical images. Qin et al. [11] used VQA models to automatically generate medical prompts containing domain-specific knowledge, enabling the application of large visual language models pretrained on natural images to medical images.

### 2.2 Visual Understanding of Disaster Images

After a disaster, extracting relevant information from images and text helped to respond to the disaster promptly. Madichetty et al. [12] proposed a new approach based on the combination of fine-tuned BERT and DenseNet [13] to extract information from Twitter tweets images and text to identify effective tweets related to disasters. Weber and Kan [14] presented the xBD satellite image dataset. The images were analyzed to assess the building damage and four damage levels were defined. The researchers improved the model performance through improved CNN and the fusion of features before the last layer of semantic segmentation.

## 3 METHODS

In this section, we give our problem definition first and then introduce our method in detail.

### 3.1 Problem Definition

Given a disaster image $I$ and a question $Q$, the goal of our method is to transform the input into a free-from answer $A$ using a generation model $Gen$. Thus, we have: $A = Gen(I, Q)$, where $Gen(\cdot)$ represents our generation model, $I$ represents the image, $Q$ represents the question, and $A$ represents the generated answer.

### 3.2 Framework

We propose the ZFDDA model with the modular idea [15] which means every module in the model is an independent module. The overall model architecture (Figure 2) consists of three modules: (1) Image Content Extraction Module, (2) Chain of Thought Demonstration, and (3) Question Answering Module. This method bridges the question-answering module with the image content extraction module with a well-designed chain of thought demonstration. Unlike supervised VQA models (e.g., the MFB model [5]), the answer space of our method is open.

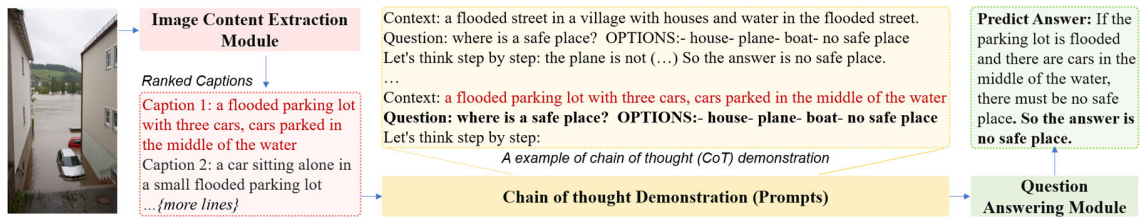

Figure 2: An overview of our ZFDDA model

#### 3.2.1 Image Content Extraction Module

**The role of this module is to output multiple image captions as candidate contexts and then select the most relevant to the question as the final context.** Firstly, captions relevant to the related regions are extracted [15]. Meanwhile, to extract more diverse information from the images, $N$ captions are generated for each image as context candidates. Finally, we calculate the similarity score between each caption and the question. The calculation method used is cosine similarity. The highest-scoring caption is selected as the final context information. It is combined with the question to be a part of a chain of thought demonstration. In addition, for the choice of $N$, we set $N$ to 5 for multiple-choice questions and 50 for the rest of the question types.

#### 3.2.2 Chain of Thought Demonstration

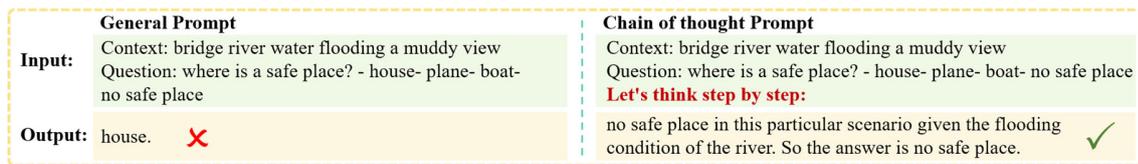

Figure 3: Difference between a general prompt and a chain of thought prompt

**CoT is used to further stimulate the potential of large language model working on the disaster VQA.** To make better use of the knowledge that language models have learned, prompts are usually added to the input of the language model. However, the general prompt has limited ability to improve the performance of tasks that require multi-step reasoning. We use another kind of prompt called the CoT prompt, where "Let's think step by step" is added at the end of the original input, and the model shows surprising and unexpected abilities when answering the question. Figure 3 demonstrates the difference between a general prompt and a chain of thought prompt.

In our method, we provide two CoT settings. One is the zero-shot CoT. The other is the few-shot CoT. The zero-shot CoT means that we only add "Let's think step by step" directly to the original input, whereas the few-shot CoT adds $M$ examples in front of the zero-shot input. These examples give the thought process so that the model can learn the details better. The differences between the two CoT settings are shown in Table 1. In our method, for multiple-choice questions, 3 examples are provided, for the remaining types of questions, we provide 1 example.

Table 1: A example of zero-shot CoT and few-shot CoT setting

| Without CoT Input | Zero-shot CoT Input | Few-shot CoT Input |
| --- | --- | --- |
| Context: [caption] Question: where is a safe place? OPTIONS: - house- plane - boat - no safe place | Context: [caption] Question: where is a safe place? OPTIONS: - house- plane - boat - no safe place **Let's think step by step:** | **Context: a flooded street in a village with houses and water in the flooded street.** **Question: where is a safe place? OPTIONS:- house- plane- boat- no safe place** **Let's think step by step: : the plane is not mentioned, the house is mentioned but the house is flooded. So the answer is no safe place.** ... Context: [caption] Question: where is a safe place? OPTIONS:- house- plane- boat- no safe place **Let's think step by step:** |

*3.2.3 Question Answering Module*

**This module outputs the final answer as well as the thought process.** The input of this module is a well-designed chain of thought demonstration. Flan-Alpaca, a pre-trained language is the main component of this module. Alpaca [16] is a language model with less training data fine-tuned from the LLaMA 7B model [17] and shorter training time, with performance comparable to that of GPT-3.5. Flan, for Finetuned Language Net [18], is the model after instruction tuning which can improve the model's performance greatly. Flan-Alpaca is an instruction-tuned model of Alpaca, which contains instruction tuning from both humans and machines.

**4 EXPERIMENTS**

In this section, we first introduce our dataset and then the baseline model used for comparison. Metric for assessment is introduced next.

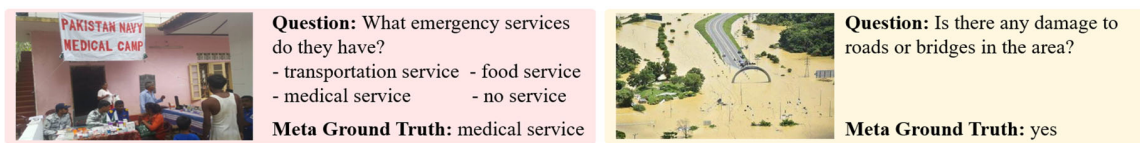

Figure 4: Some example questions and their corresponding images and meta ground truths.

**4.1 Dataset**

**FFD-IQA:** To comprehensively evaluate our method, we construct a large-scale FFD-IQA dataset. Our dataset includes 2,058 images and 22,422 question-meta ground truth pairs focusing on the safety of individuals trapped in disaster sites and the availability of emergency services. The images are collected from the CrisisMMD [20], FloodNet [19], and European Flood 2013 Dataset [21]. There are three types of questions are included: *Free-form* questions, *Multiple-choice* questions, and *Yes-no* questions. *Yes-no* questions such as "Is there any elderly person in the area?" concentrate on whether there are vulnerable people who are harmed and are about to be harmed. The *Multiple-choice* questions such as "Where is a safe place? -house -plane-boat-no safe place" help affected populations quickly locate nearby safe shelters and get emergency services. *Free-form* questions don't have a specific format which means we could bring questions about any content in the photo. Dataset examples are in Figure 4.

**Automatic Dataset Construction Pipeline:** This is a three-phase pipeline. The pipeline consists of three modules: *image content extraction module*, *object matching module*, and *question generation module*. Given an image, firstly, the descriptions of the image and the main entities are extracted; after that, we use a visual ground method [22] to check the image contains all the objects; finally, the entity nouns are taken as the Meta Ground Truth to generate the question using the question generation model [23].

### 4.2 Baseline

**Plug-and-Play VQA [15]:** Plug-and-Play VQA is a zero-shot VQA model without architectural modifications and additional training to any module. It achieves the task of answering questions by connecting a pre-trained image caption model and a pre-trained language model. Compared to previous attempts at zero-shot VQA models, Plug-and-Play VQA saves training time and achieves better experimental results on both VQAv2 [24] and GQA [25] datasets. As an excellent VQA model, we use the experimental results as the baseline for our model.

### 4.3 Metric

The general metrics to assess the quality of generative VQA models are BLEU score, CIDEr [26], etc. These metrics are not suitable for our task because our question does not have a standard answer. We evaluate the experiment results through manual evaluation. Based on the scoring criteria introduced below, we invite 5 evaluators to ask them to rate the output of each method. A plausible answer is evaluated as a score of "1" and an implausible answer is evaluated as a score of "0". Accuracy is our evaluation metric which is defined as the ratio of the number of "1" to the number of total questions in this paper. Formally, the Accuracy $A$ can be calculated as $A = \frac{count(p)}{total(q)}$, where $p$ refers to a question with the plausible answer, $q$ refers to the question, $count(p)$ refers to the total number of questions with the plausible answer, and $total(q)$ refers to the total number of questions.

**Human Evaluation:** To avoid subjectivity in evaluator assessment and inconsistency in evaluation criteria, when selecting evaluators for manual evaluation, this paper strictly adheres to the Inter Annotator Agreement (IAA) [27] during the process. We use Fleiss' Kappa as an indicator of consistency between evaluators. According to Fleiss' Kappa, we select the evaluators with the best consistency (Fleiss' Kappa=0.72) as evaluators.

**Scoring Criteria:** The evaluator rates the quality of the output of the answer by the models based on the scoring criteria. Also, we provide the description of plausible and implausible answers below:

- **Implausible answer:** For a method without CoT, an implausible example is that the answer does not match the content of the image, e.g., there are no vehicles in the image, but there is a transportation service in the answer. Another implausible example is for models that introduce CoT, the answer is correct but the reasoning process does not match the content of the image. For example, the answer is yes for road damage, but the reasoning process says that no damage has been done to roads or bridges in the area.
- **Plausible answer:** A plausible example is that the answer does match the image. The inference process for the model that introduces CoT also matches the image. For example, if the image shows the road is flooded, it is plausible that the model reasons that there is no safe place.

## 5 EXPERIMENT RESULTS

In this section, we conduct experiments to analyze the effectiveness of our methods. Then we analyze the results of the ablation experiments to verify the effectiveness of introducing the CoT prompt.

Table 2: Accuracy on different methods

| Methods | All | Multiple-choice | Free-form | Yes-no |
|---|---|---|---|---|
| PNP-VQA | 52.07% | 29.45% | 42.62% | 61.09% |
| ZFDDA w/o CoT | 52.06% | 32.05% | 62.18% | 55.03% |
| ZFDDA zero-shot CoT | 57.43% | 33.21% | 83.26% | 57.36% |
| ZFDDA few-shot CoT | 64.80% | 57.00% | 86.13% | 61.25% |

### 5.1 Performance Analysis of ZFDDA Model

Table 2 shows the performance of ZFDDA as well as other methods on the FFD-IQA dataset. According to Table 2, the following conclusions can be drawn: (1) Under without CoT prompts setting, the accuracy of our ZFDDA w/o CoT method improved by 19.56% in answering free-form questions, from 42.62% to 62.18% when compared to the PNP-VQA model. Our ZFDDA w/o CoT model is more accurate in answering multiple-choice and free-form questions. The performance of our model is also competitive on yes-no type of questions. (2) Under the zero-shot CoT prompts setting, the accuracy of our ZFDDA zero-shot CoT method improved by 3.76% in answering multiple-choice questions, from 29.45% to 33.21% when compared to the PNP-VQA model. So the introduction of CoT is of vital importance. (3) Also, the weak differences in answering yes-no questions between our method and the baseline may also stem from whether the question-answer generation module is rich in context information as input. In the PNP approach, 20 captions are combined as context information. In our approach, we only select the most relevant caption as context information. How to generate a comprehensive caption that contains the detailed content of the image is an important task for us to improve in the following research.

### 5.2 Ablation Studies

**Effectiveness of Applying CoT:** The introduction of CoT increases the accuracy in answering questions for all types of questions. As shown in the experimental results, the chain of thought demonstrations can substantially improve the performance of our method. For example, according to Table 2, for the free-form questions, the ZFDDA zero-shot CoT method improves the accuracy by 21.08%, from 62.18% to 83.26%.

**Effectiveness of examples in Few-shot CoT:** Compared with the ZFFDA zero-shot CoT method, the ZFFDA few-shot CoT model improves the accuracy further. As shown in Table 2, for the multiple-choice questions, the ZFDDA few-shot CoT model improves the accuracy by 23.79%, from 33.21% to 57.00%. The examples in the few-shot CoT provide a thought process for the language model. They are necessary for answering the more complex multiple-choice questions.

## 6 CONCLUSION AND FUTURE WORK

In this paper, we propose an intelligent VQA model for flood disaster damage assessment by drawing on the modularity idea, which bridges the blank of the application of the zero-shot VQA model in disaster scenarios. In the model, by adding the chain of thought demonstration we inspire the knowledge transfer ability and reasoning ability of the model so that the model trained in the natural domain can be used in the domain of disaster under the zero-shot setting. To evaluate our model, we build a dataset of image question-meta ground truth pairs about flood disasters. After experiments, our model expresses high competitiveness in our dataset. At current our model is not reaching optimal performance. In the future, we will investigate the generalization of the model and its accuracy in other disaster scenarios.

**ACKNOWLEDGMENTS**

This work was supported by the Program of Natural Science Foundation of Shanghai (No. 23ZR1422800).